\documentclass[11pt]{article}
\usepackage{tikz}
\usepackage{pgf-pie}
\usepackage{xcolor}
\usepackage{acl}
\usepackage[utf8]{inputenc}
\usepackage{times}
\usepackage{latexsym}
\usepackage{multirow}
\usepackage{booktabs}
\usepackage{amsmath}

\usepackage[T1]{fontenc}

\usepackage{microtype}

\title{MITRA: A Large-Scale Parallel Corpus and Multilingual Pretrained Language Model for Machine Translation and Semantic Retrieval for Pāli, Sanskrit, Buddhist Chinese, and Tibetan}

\author{
  {\bf Sebastian Nehrdich}$^{1}$ \quad
  {\bf Kurt Keutzer}$^2$
  \\
  \\
  $^1$Center for Integrated Japanese Studies, Tohoku University,\\
  $^2$University of California, Berkeley, Berkeley Artificial Intelligence Research (BAIR) \\
}
\begin{document}
\maketitle

\begin{abstract}
Ancient Buddhist literature features frequent, yet often unannotated, textual parallels spread across diverse languages: Sanskrit, Pāli, Buddhist Chinese, Tibetan, and more. The scale of this material makes manual examination prohibitive. We present the MITRA framework, which consists of a novel pipeline for multilingual parallel passage mining, MITRA-parallel, a large-scale corpus of 1.74 million parallel sentence pairs between Sanskrit, Chinese, and Tibetan, and the development of the domain-specific pretrained language model \texttt{Gemma 2 MITRA}. We present \texttt{Gemma 2 MITRA-MT}, a version of this base model fine-tuned on machine translation tasks, reaching state-of-the-art performance for machine translation of these languages into English and outperforming even much larger open-source models. We also present \texttt{Gemma 2 MITRA-E}, a semantic embedding model that shows state-of-the-art performance on a novel, detailed semantic embedding benchmark. We make the parallel dataset, model weights, and semantic similarity benchmark openly available to aid both NLP research and philological studies in Buddhist and classical Asian literature.
\end{abstract}

\section{Introduction}
Over the course of more than two millennia, the Buddhist tradition has produced a massive body of literature nowadays preserved primarily in Pāli, Sanskrit, Buddhist Chinese, and Tibetan. Within this literature, semantically related or closely matching passages of variable length are frequently encountered. These parallels appear both within literature preserved in the same language and across different languages. Multilingual parallelism is especially prominent due to large-scale translation efforts of Indic Buddhist text material by the Buddhist traditions: first into Chinese (beginning in the 2nd century CE), and later into Tibetan (from the 8th century CE onwards). Translation of this body of literature into modern English is ongoing, but so far only a fraction of the material has been properly translated. To give one example, about 10\% of the digitally available Buddhist texts in Chinese have been translated into English so far~\cite{nehrdich2023}.\\
Textual reuse within Buddhist literature, while occurring frequently, is often not marked explicitly and is therefore a significant research objective for philologists~\cite{freschi2014reuse}. Exhaustive manual studies remain prohibitively labor-intensive. Consequently, existing research is limited to specific topics within subsections of the literature (for one example, see~\citet{hellwig-etal-2023-vedic}). Since the Buddhist textual transmission is highly multilingual, parallel data is needed for the training of efficient multilingual retrieval models, which is currently only exhaustively collected between Sanskrit and Tibetan~\cite{nehrdich2022}.\\
In order to address these challenges, we propose a framework that utilizes machine translation into English as a pivot step to retrieve alignment candidates of longer passages, which we then further refine with sentence-level alignment. We then combine the generated multilingual parallel data with other monolingual and parallel resources in order to pretrain a domain-specific large language model (LLM) for these languages. We then fine-tune two different versions of this model: one for machine translation and one for multilingual semantic retrieval. Our benchmarks show that this LLM outperforms all other open baselines on machine translation and semantic retrieval for these languages.\\
This paper makes the following contributions: 
\begin{itemize}
\item The description of a pipeline for the retrieval of matching parallel passages within literature preserved in different classical Asian languages based on machine translation as a pivot step
\item MITRA-parallel, a novel dataset of documents automatically aligned at the sentence level across Sanskrit, Chinese, and Tibetan, comprising 1,742,786 sentence pairs
\item Description of the continuous pretraining of \texttt{Gemma 2 MITRA}, a domain-specific Gemma 2-based large language model
\item Fine-tuning of this model on both machine translation and information retrieval tasks
\item Evaluation of the machine translation performance for Sanskrit, Pāli, Tibetan, and Chinese-to-English tasks
\item A novel benchmark for multilingual semantic retrieval for ancient Buddhist languages with a dataset covering seven different tasks
\item Comprehensive evaluation of different retrieval methods using a unified testing protocol on this evaluation dataset and an ablation study for monolingual retrieval methods
\end{itemize}

We provide the automatically generated novel dataset of sentence-level aligned ancient Buddhist document pairs, evaluation dataset, testing protocol and links to the trained model weights at \url{https://github.com/dharmamitra/mitra-parallel}.\\
Additionally, we provide access to this dataset through a user-friendly, searchable online database designed for philological research in Buddhist and broader classical Asian literature at \url{https://dharmanexus.org}. 

This paper proceeds as follows: Section~\ref{sec:previous-research} introduces relevant previous work. Section~\ref{section:dataset} describes the data mining procedure used to create MITRA-parallel, and presents statistics and a quality assessment of this dataset. Section~\ref{section:gemma2-mitra-base-llm} describes the continuous pretraining of the \texttt{Gemma 2 MITRA} base language model, including the composition of its multilingual pretraining data, and the instruction finetuning processes for both machine translation and semantic retrieval tasks. Section~\ref{section:mt-evaluation} presents the evaluation of the machine translation capabilities of \texttt{Gemma 2 MITRA-MT}, comparing its performance on translating Sanskrit, Pāli, Tibetan, and Buddhist Chinese into English against other open large language models and existing domain-specific models. Section~\ref{section:retrieval-evaluation} introduces a novel benchmark for multilingual semantic retrieval and evaluates the performance of \texttt{Gemma 2 MITRA-E} against various sparse and dense retrieval methods across four distinct retrieval scenarios. Section~\ref{section:ablation-study} conducts an ablation study to assess the monolingual retrieval performance of \texttt{Gemma 2 MITRA-E} compared to other models on Sanskrit and Chinese texts. Section~\ref{section:conclusion} summarizes our contributions and discusses potential future work, while Section~\ref{section:limitations} describes the limitations of this study.

\section{Previous Research}\label{sec:previous-research}
Recent scholarship has increasingly employed computational methods to detect textual parallels within ancient Buddhist literature. Efforts include identifying reuse in Sanskrit corpora \citep{hellwig-googling-rsi}, Tibetan texts using string similarity \citep{klein2014}, and Buddhist Chinese literature via word embeddings and alignment algorithms \citep{nehrdich2020}.\\
Cross-lingual investigations have explored aligning Tibetan and Buddhist Chinese sentences with static embeddings \citep{felbur2022} and matching Sanskrit and Tibetan documents using deep neural transformer-based sentence representations \citep{nehrdich2022}. Despite these advances, there is a lack of large-scale, sentence-aligned multilingual resources spanning Sanskrit, Pāli, Buddhist Chinese, and Tibetan.\\
While deep neural sentence representations are state-of-the-art for semantic similarity \citep{reimers2019} and bitext mining \citep{schwenk-2018-filtering, artetxe-schwenk-2019-margin}, with powerful multilingual models available for high-resource languages \citep{laser2019, labse2022}, the application of this approach to this specific multilingual ancient setting is not yet explored. Similarly, powerful dense retrieval models \citep{karpukhin-etal-2020-dense}, crucial for tasks like RAG \citep{rag2020}, often lack specialization for these historically significant, low-resource languages. Our work addresses these gaps by developing a comprehensive parallel corpus and domain-specific models.

\section{MITRA-parallel dataset}\label{section:dataset}
Ancient Buddhist literature features extensive multilingual parallels, primarily from Indic texts translated into Chinese and Tibetan. However, these are often uncataloged and may exist only as fragments (e.g., chapters or paragraphs within larger works), making identification challenging. We define our task as detecting common sub-passages of semantically equivalent parallel sentences of at least paragraph-length. The highly repetitive nature of Buddhist texts means standard sentence-level mining approaches \citep{schwenk-2018-filtering, artetxe-schwenk-2019-margin} are prone to generating excessive noise, thus requiring a more constrained mining pipeline.
  
\subsection{Data Mining Procedure}
Our pipeline leverages the tendency of parallel sentences to appear in continuous chains. It proceeds in three main stages:
\paragraph{Machine Translation} 
All documents are translated into English using a MADLAD-400 model \citep{madlad} fine-tuned on domain-specific data \citep{nehrdich2023}, including 2 million Tibetan-English pairs (monlam.ai) and a forthcoming Sanskrit-English dataset.
\paragraph{Candidate Clusters}
On these English translations, we generate overlapping sliding windows (concatenated adjacent sentences to a minimum length for higher retrieval precision). These windows are embedded using BGE M3 \citep{bgem3}. Corpus-wide kNN search on cosine similarity identifies initial candidate pairs $(x_i, y_i)$ (source/target text positions). Spatial hashing then efficiently groups these pairs into clusters $C_k = {C_1, C_2, \dots }$, each representing a contiguous region of likely parallelism.
\paragraph{Sentence Alignment}
The identified candidate regions are refined to precise sentence-level alignments using \textsc{BERTAlign} \citep{bertalign} with a domain-finetuned \texttt{LaBSE} model \citep{labse2022}. Crucially, this alignment operates on the original language sentences, not the translations, helping to remove noisy matches at cluster peripheries. A final filtering step applies a moving average threshold on the sentence pairs' average cosine similarity to ensure quality.\\\\
This three-step process consisting of machine translation, coarse region identification, and fine-grained sentence alignment, significantly reduces noise from the repetitive nature of Buddhist literature, yielding high-quality parallels suitable for both machine learning and philological research.

\subsection{Generated Dataset}\label{section:dataset-desc}
\begin{figure}[htbp]
    \centering
    \includegraphics[width=0.45\textwidth]{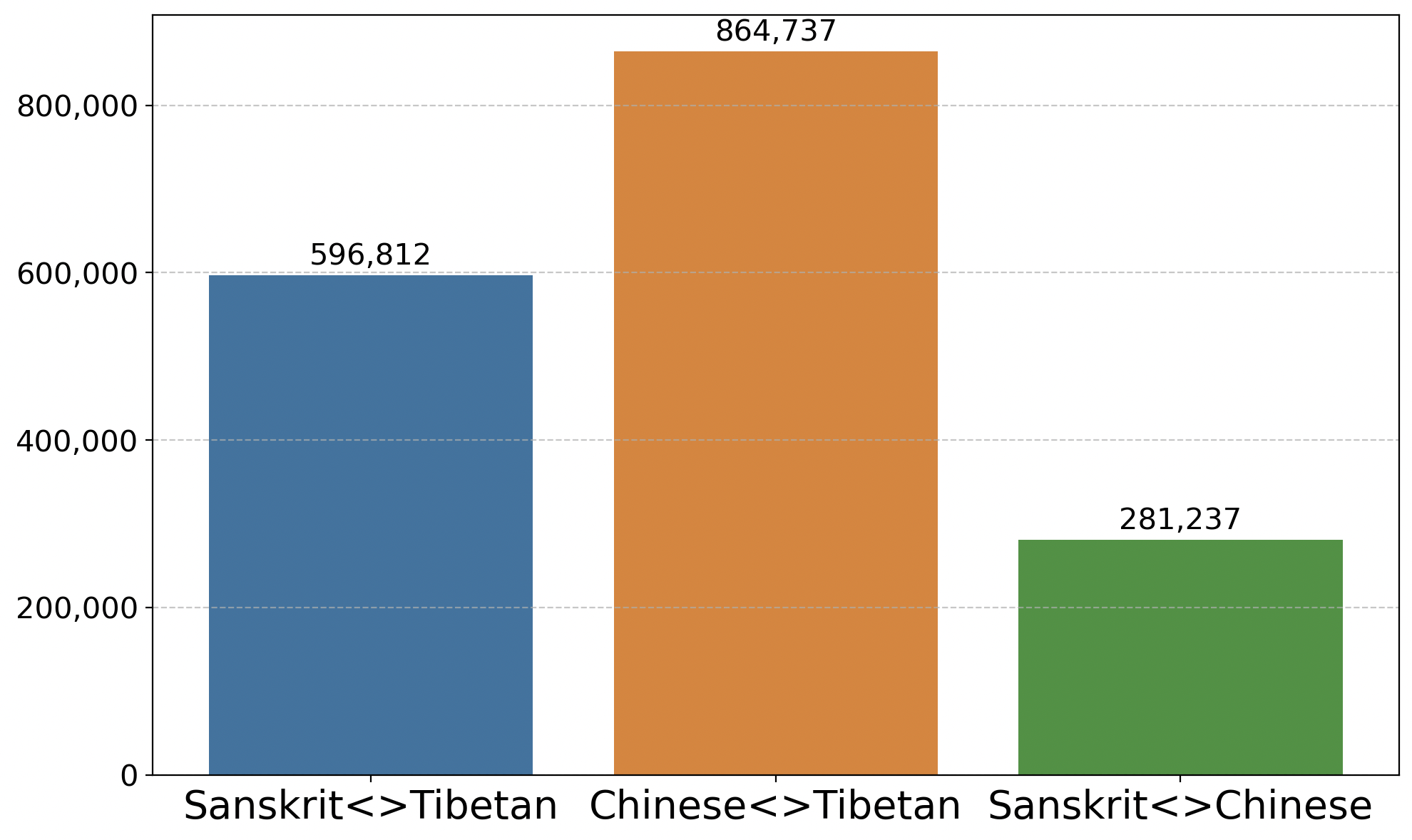}
    \caption{Number of datapoints per language pair in the sentence-level aligned parallel dataset for Ancient Buddhist Languages. }
    \label{fig:buddhist-languages-dataset}
\end{figure}

This method yielded 1,742,786 parallel sentence pairs across Sanskrit<>Tibetan, Chinese<>Tibetan, and Sanskrit<>Chinese (see Figure \ref{fig:buddhist-languages-dataset} for an overview). This significantly expands existing resources. For instance, our 596,812 Sanskrit<>Tibetan pairs represent an 89\% increase over the SansTib dataset \citep{nehrdich2022}. For Sanskrit<>Chinese and Chinese<>Tibetan, this work provides the first large-scale dedicated digital parallel datasets.

To assess quality, we manually evaluated 100 randomly sampled pairs, categorizing them as 'Correct' (perfect match), 'Partially Correct' (majority overlap with minor mismatch), or 'Wrong' (<50\% correct correspondence). As shown in Table \ref{tab:error_rates}, 73\% achieved perfect alignment, indicating a strong baseline for unsupervised retrieval. While 11\% were 'Wrong', some may still correctly identify document pairs with the help of such matches, and deterministic filters (e.g., segment length ratios) can remove such misalignments effectively for ML applications.

\begin{table}[h]
\centering
\begin{tabular}{|l|c|}
\hline
\textbf{Category} & \textbf{Percentage (\%)} \\
\hline
Perfect & 73 \\
\hline
Partly correct & 16 \\
\hline
Wrong & 11 \\
\hline
\end{tabular}
\caption{Manual examination of error rates in the mined dataset. Perfect means that both source and target segment match up without any errors. Partly correct means that more than 50\% of source and target segment match up. Wrong means that less than 50\% of source and target segment match.}
\label{tab:error_rates}
\end{table}


\section{Gemma 2 MITRABase LLM}\label{section:gemma2-mitra-base-llm}
We pretrain our own language model on a large dataset of relevant Buddhist data which serves as the basis for the machine translation and semantic retrieval model. We follow the training recipe of \textsc{Tower}~\cite{tower2024}: We take a strong baseline model, in this case Gemma~2~\cite{gemma2}, and continuously pretrain it on a large domain-specific multilingual corpus that consists of both monolingual data in all four languages as well as high-quality parallel data. We decided to build on top of Gemma~2 since in our evaluation, it has shown the most consistent baseline performance on machine translation into English compared to other open LLMs such as Llama~3, Mistral, or the Qwen family (see Section~\ref{section:mt-evaluation}). Our continuous pretraining dataset consists of a total number of 4.4 billion tokens.
\paragraph{Monolingual data} 40\% of the data is domain-specific English data consisting of academic works on Buddhist Studies and related disciplines, as well as translations of Buddhist texts into English acquired from various academic sources, which we processed with Google Cloud Vision OCR. We applied simple rule-based cleaning to remove lines that consist largely of non-alphabetic characters. We deduplicate the dataset on the document level. 20\% of the data is Sanskrit and Pāli data collected from various digital resources of Sanskrit and Pāli texts. We do not use any non-corrected OCR Indic material directly. 15\% is Buddhist Chinese from the CBETA collection\footnote{\url{https://github.com/cbeta-org/xml-p5}} and 5\% is Tibetan sourced via the Asian Classics Input Project (ACIP).

\paragraph{Parallel data} 20\% of the data consists of multilingual parallel sentence pairs. The basis of the dataset are the mined sentence pairs described in Section~\ref{section:dataset-desc}. We further collected 1M sentence pairs between Sanskrit and English (publication under preparation). We added 2M sentences between Tibetan and English sourced via our collaborative effort with monlam.ai. The Kumarajiva project\footnote{\url{https://khyentsefoundation.org/kf-projects/kumarajiva-project/}} contributed 41,000 gold-quality Tibetan$\leftrightarrow$Buddhist Chinese sentence pairs. We also collected 31,000 gold-quality Sanskrit$\leftrightarrow$Buddhist Chinese sentences. We further use 149,418 Pāli-to-English sentence pairs. 

\paragraph{Continuous Pretraining} We train the Gemma~2 model (without instruction fine-tuning) with a size of 9B parameters for two epochs on this dataset. We use an effective batch size of 2M tokens per gradient step. We set the maximum length at 1024 tokens. We used the DeepSpeed library\footnote{\url{https://www.deepspeed.ai}} with ZeRO Stage~3 for half-precision training in fp16. The pretraining took four weeks on 8$\times$ A100. We refer to this continuously fine-tuned version of Gemma~2 as \texttt{Gemma~2 MITRA} in this paper.
\paragraph{Machine Translation Instruction Fine-tuning} 
We use the Claude 3.5 Sonnet API (September 2024) in order to mine 10,000 multi-direction translation examples and 30,979 document-level Sanskrit/Pāli/Tibetan/Chinese-to-English examples. We fine-tuned the model for four epochs on this dataset. In our examination, mining instruction data from high-performing LLMs leads to preferable results over using gold-quality human created sentence pairs, which lead to frequent repetitive hallucinations when used for instruction fine-tuning, which occur much less frequently when using LLM-generated instruction data. 
\paragraph{Semantic Retrieval Fine-tuning} 
We fine-tune \texttt{Gemma 2 MITRA} for semantic retrieval using contrastive loss with task-specific prompts~\citep{bgeicl2024}. Original data for these tasks is scarce, so we augmented it using the Gemini 2.0 Flash API (March 2025). The fine-tuning dataset is detailed in Table~\ref{tab:dataset-stats}. In this table, the first two blocks describe retrieval tasks where the goal is to find a matching translation sentence in the target language given a source sentence. Regarding the various tasks, \textbf{Keywords (eng)} describes the retrieval of a passage in any of the four languages based on a number of English keywords. \textbf{Keywords} describes the same task but with keywords in the same language, i.e., Sanskrit keywords for the retrieval of a Sanskrit sentence. \textbf{Questions (eng)} describes the retrieval of a passage in any of the four languages based on a question in English. \textbf{Summary (eng)} is the same task, but based on an English summary rather than a question. \textbf{Sentence} describes the retrieval of a larger section based on a single, short sentence in the same language.
\begin{table}[t]
\centering
\small
\begin{tabular}{l|l|l|l|l}
\toprule
\textbf{Task} & \textbf{Source} & \textbf{Target} & \textbf{Type} & \textbf{Samples} \\
\midrule
\multirow{4}{*}{English$\rightarrow$X} & English & Chinese & Orig & 50,000 \\
& English & Pali & Orig & 50,000 \\
& English & Sanskrit & Orig & 50,000 \\
& English & Tibetan & Orig & 50,000 \\
\midrule
\multirow{7}{*}{Multilingual} & Pali & Chinese & Orig & 4,809 \\
& Pali & Pali & Orig & 247 \\
& Pali & Sanskrit & Orig & 4,613 \\
& Pali & Tibetan & Orig & 11,184 \\
& Sanskrit & Chinese & Orig & 50,000 \\
& Sanskrit & Tibetan & Orig & 50,000 \\
& Tibetan & Chinese & Orig & 50,000 \\
\midrule
\multirow{5}{*}{Various} & \multicolumn{2}{l|}{Keywords (eng)} & Synth & 47,223 \\
& \multicolumn{2}{l|}{Keywords} & Synth & 47,997 \\
& \multicolumn{2}{l|}{Questions (eng)} & Synth & 43,321 \\
& \multicolumn{2}{l|}{Summary (eng)} & Synth & 38,882 \\
& \multicolumn{2}{l|}{Sentence} & Synth & 51,382 \\

\bottomrule
\end{tabular}
\caption{Instruction finetuning dataset for semantic retrieval. Type "Orig" describes original data, while "Synth" describes synthetic data.}
\label{tab:dataset-stats}
\end{table}
\section{Machine Translation Evaluation}
\label{section:mt-evaluation}
We use manually selected sentence pairs for evaluation of machine translation quality into English for each language: 2,662 sentence pairs of Buddhist Chinese$\leftrightarrow$English taken from~\citet{nehrdich2025}. For Sanskrit, we use a total of 5,552 sentence pairs selected from a number of domains, including Buddhist Sūtras as well as other domains such as Vedic ritual and poetry (publication under preparation). For Tibetan, we use 4,053 sentence pairs randomly sampled from the entirety of the sentence pair data. Since the Pāli canon is heavily dominated by the Sutta collection with its repetitive language, we sampled 1,900 sentence pairs of mostly non-canonical material (\textit{Jātakagāthāvaṇṇanā}, \textit{Navapadamañjarī}, and \textit{Cariyāpiṭaka}) for which domain-wise very little intersection with our existing training corpus exists. All evaluation data points have been removed from the pretraining/fine-tuning stages of  \texttt{Gemma~2 MITRA}. While \texttt{Gemma~2 MITRA-MT} was fine-tuned on translation into a number of different languages, we limit our evaluation here to translation into English.

We compare the following models: Mistral 7B v0.3 IT~\cite{mistral}, Llama 3.1 8B Instruct~\cite{llama3}, Qwen2.5 7B~\cite{qwen25}, Gemma~2 9B IT~\cite{gemma2}, and Gemma 3 in the 12B IT and 27B IT variants~\cite{gemma3}.

\begin{figure*}[htbp]
    \centering
    \includegraphics[width=0.95\textwidth]{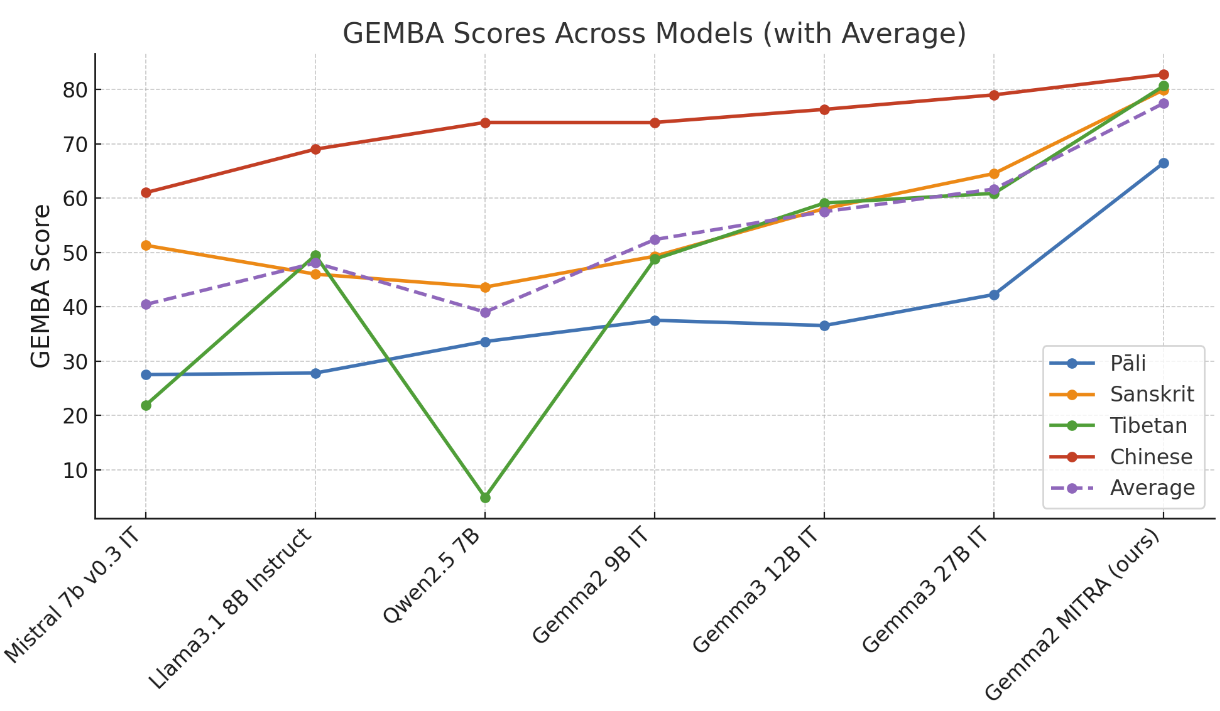}
    \caption{Machine translation performance of different open base models compared to our finetuned model. Performance is measure in GEMBA score, which we implemented with Gemini 2.0 Flash as judge.}
    \label{fig:mt-performance}
\end{figure*}

We present the results of the machine translation evaluation of Gemma~2 MITRAagainst other open models in Figure~\ref{fig:mt-performance}. We use GEMBA with Gemini 2.0 Flash as evaluation metric due to its strong performance on ancient Asian languages~\cite{nehrdich2025}. The results show that \texttt{Gemma~2 MITRA-MT} outperforms all other open LLMs by a significant margin. All models do best on Buddhist Chinese, which indicates that transfer learning from a closely related high-resource language pair, Modern Chinese and English in this case, strongly benefits this idiom. Mistral 7B IT is comparatively strong on Sanskrit, but struggles with Tibetan and Pāli. Llama 3.1 8B Instruct performs rather well on Tibetan, but falls slightly behind on the other languages. Qwen2.5 7B performs comparatively well on Chinese and Pāli, but shows extremely weak performance on Tibetan. Gemma 2 9B IT is very consistent across all four languages. Gemma 3 12B IT further outperforms Gemma 2 9B IT, showing consistent performance increases over time in the Gemma family. Gemma 3 27B IT further outperforms the 12B version on all languages, indicating that larger parameter count does lead to better MT performance. \texttt{Gemma~2 MITRA-MT} outperforms Gemma 3 27B IT by a significant margin for all languages. While the performance of Sanskrit, Tibetan, and Buddhist Chinese converge on a similar plateau after fine-tuning, Pāli falls behind. The likely reason for this is that our evaluation data heavily relies on commentarial material, for which very little has been translated into English at all, and it is therefore heavily underrepresented in the training data.

We also evaluate the performance of \texttt{Gemma 2 MITRA-MT} against the only other domain-specific open-source model \texttt{MITRA NMT ZH-EN}\footnote{\url{https://huggingface.co/buddhist-nlp/mitra-mnt-zh-en}} for Buddhist Chinese~\cite{nehrdich2023}. \texttt{MITRA NMT ZH-EN} is a model based on Facebook AI's 2021 WMT submission~\cite{tran-etal-2021-facebook} with further domain-specific fine-tuning. We present the results in Table~\ref{tab:zh_en_translation_results}. In this setting, we also evaluate on chrF~\cite{popovic-2017-chrf} and BLEURT~\cite{bleurt} scores in addition to GEMBA. \texttt{Gemma 2 MITRA-MT} outperforms \texttt{MITRA NMT ZH-EN} on all metrics, establishing a new state-of-the-art for open models on Buddhist Chinese-to-English machine translation among open models.
\begin{table}[t]
\centering
\small
\begin{tabular}{lccc}
\toprule
\textbf{Model} & \textbf{chrF} & \textbf{BLEURT} & \textbf{GEMBA} \\
\midrule
MITRA NMT ZH-EN & 32.14 & 0.551 & 67.41 \\
Gemma 2 X-MT & \textbf{36.59} & \textbf{0.579} & \textbf{82.78} \\
\bottomrule
\end{tabular}
\caption{Chinese-English translation performance on the MITRA ZH-eval test set. M}
\label{tab:zh_en_translation_results}
\end{table}

\section{Retrieval System Evaluation} \label{section:retrieval-evaluation}
Our evaluation framework assesses cross-lingual and monolingual passage retrieval across four distinct scenarios:
\paragraph{Modern English -> Classical Retrieval} Retrieving matching Sanskrit, Tibetan, Chinese, and Pāli segments using contemporary English queries. Data consists of manually verified English translations paired with original classical sentences, analogous to the BUCC mining task \citep{bucc2017,bucc2018}.
\paragraph{Cross-lingual Parallel Retrieval}
Evaluating precision for retrieving parallel Classical Buddhist text segments across Sanskrit-Tibetan, Sanskrit-Chinese, and Chinese-Tibetan pairs. This manually verified data is also analogous to BUCC.
\paragraph{Verse $\rightarrow$ Commentary Retrieval} Specialized tasks involving retrieving commentary passages from root texts. We test four cases: Sanskrit root $\rightarrow$ Sanskrit commentary, Chinese root $\rightarrow$ Chinese commentary, Tibetan root $\rightarrow$ Tibetan commentary, and the cross-lingual Sanskrit root $\rightarrow$ Tibetan commentary. These present challenges due to variable passage relatedness, length, and content imbalances.
\paragraph{Cross-lingual QA Retrieval} Retrieving answers in Classical languages (Pāli, Sanskrit, Tibetan, Chinese) to English questions.
\paragraph{Evaluation Protocol} In order to simulate a real-world retrieval scenario where the right data point needs to be retrieved from a large search space of potentially millions of sentences, we randomly sample a total of 400,412 sentences from the Pāli, Sanskrit, Tibetan, and Chinese target corpora, giving each corpus 25\% weight. These sentences are used as negatives in the retrieval setup. We decided on this number of hard negatives to strike a balance between being close to the real-world retrieval application and at the same time making the evaluation, where multiple tasks need to be evaluated in multiple different setups, feasible in reasonable time.\\
We compare both sparse and dense information retrieval approaches. As sparse method we evaluate \texttt{BM25} on the English pivot machine translation. We implement \texttt{BM25} via the rank-bm25 Python library utilizing the \texttt{BM25Okapi} algorithm. We evaluate the \texttt{BGE-M3} embeddings on the English pivot in two configurations: \texttt{BGE-M3 base}, where we apply the model as it is, and \texttt{BGE-M3 ft}, where we fine-tune it on 100k samples of English-to-English pairs achieved by machine-translating source and target of random sentence pairs of the MITRA-parallel dataset described in Section~\ref{section:dataset} into English. We evaluate LaBSE directly on the input languages without English pivot, after fine-tuning it on domain-specific parallel sentence data taken from MITRA-parallel. In the same way we evaluate \texttt{Gemma 2 MITRA-E} directly on the input languages without English pivot.

\paragraph{Results} We present the evaluation results in Table~\ref{tab:retrieval-results}. \texttt{Gemma 2 MITRA-E} outperforms all other models by a significant margin on all tasks, the only exception being the P@1 accuracy on the Sanskrit verse to Tibetan commentary retrieval task, where it is outperformed by \texttt{BGE-M3 base}. \texttt{BGE-M3 ft} outperforms \texttt{BGE-M3 base} consistently, demonstrating that the domain-specific fine-tuning of this model yields considerable performance improvements even when pivoting through a different language. \texttt{LaBSE}, which was fine-tuned only on parallel sentence pair data, performs best on the modern English$\rightarrow$classical and cross-lingual parallel retrieval tasks, while lagging behind \texttt{BGE-M3} and \texttt{Gemma 2 MITRA-E}. The gap is more pronounced for the more semantically distant tasks like verse$\rightarrow$commentary and cross-lingual QA retrieval. \texttt{BM25} with English as pivot is outperformed by \texttt{LaBSE} on the first two tasks, but on the commentary and QA retrieval tasks their performance is very comparable. The results show that machine translation as a pivot step in combination with a versatile embedding model like \texttt{BGE-M3} is an improvement over native multilingual embedding systems such as \texttt{LaBSE} that are primarily trained on parallel data. While \texttt{BM25} on the machine translation yields results somewhat comparable to \texttt{LaBSE}, it is outclassed by \texttt{BGE-M3} in all cases, which makes it the preferable model for this data, especially after fine-tuning on domain-specific English data. \texttt{Gemma 2 MITRA-E} significantly outclassing all other models demonstrates that the combination of monolingual and parallel pretraining data, comparatively high parameter count, and task-appropriate instruction tuning creates a model that adapts well to all given tasks.

\begin{table*}[t]
\centering
\scriptsize
\begin{tabular}{l|l|l|c|c|c|c|c}
\toprule
\textbf{Task Type} & \textbf{Source} & \textbf{Target} & \textbf{BM25} & \textbf{LaBSE} & \multicolumn{2}{c|}{\textbf{BGE M3 (MT)}} & \textbf{Gemma 2 MITRA-E} \\
\cmidrule(lr){6-7} 
 & & & \textbf{(MT)} & \textbf{ft} & \textbf{base} & \textbf{ft} & \textbf{ft} \\
\midrule
\multirow{4}{*}{\shortstack[l]{Modern-English->\\Classical\\Retrieval}}
& English & Sanskrit & 33·45·50 & 48·64·70 & 74·82·85 & 84·90·92 & \textbf{95·98·99} \\
& English & Tibetan & 38·50·55 & 73·85·88 & 68·79·82 & 76·86·89 & \textbf{95·98·99} \\
& English & Chinese & 23·36·40 & 53·68·73 & 58·70·74 & 69·79·83 & \textbf{90·95·96} \\
& English & Pāli & 28·43·48 &  30·52·59 & 53·70·75 & 60·76·81 & \textbf{86·97·98} \\
\midrule
\multirow{3}{*}{\shortstack[l]{Cross-lingual\\Parallel\\Retrieval}} 
& Sanskrit & Tibetan & 42·57·62 & 54·69·74 & 69·82·85 & 77·88·91 & \textbf{93·98·98} \\
& Sanskrit & Chinese & 14·23·28 &  19·33·39 & 29·45·51 & 40·59·65 & \textbf{79·94·96} \\
& Chinese & Tibetan & 17·27·31 & 32·50·57 & 36·54·58 & 46·63·68 & \textbf{72·85·87} \\
\midrule
\multirow{6}{*}{\shortstack[l]{Sanskrit Verse->\\Sanskrit Commentary\\ Retrieval}}
& BGh & BGh & 29·43·48 & 31·40·45 & 53·61·63 & 60·70·74 & \textbf{89·95·96} \\
& MSABh & MSABh & 05·18·22 & 07·21·26 & 11·28·34 & 11·36·44 & \textbf{14·64·70} \\
& DrāhŚS & DrāhŚS & 11·16·18 & 06·10·12 & 11·16·19 & 14·21·24 & \textbf{37·50·53} \\
& LāṭŚS & LāṭŚS & 24·35·40 & 19·27·31 & 31·42·46 & 34·44·49 & \textbf{54·64·67} \\
& ŚāṅkhŚS & ŚāṅkhŚS & 14·22·25 & 10·15·18 & 18·26·29 & 20·27·30 & \textbf{50·65·69} \\
& JaimŚS & JaimŚS & 11·18·21 & 07·11·12 & 13·21·23 & 12·21·23 & \textbf{35·52·58} \\
\midrule
\multirow{3}{*}{\shortstack[l]{Chinese Verse->\\Chinese Commentary\\ Retrieval}}
& T1552 & T1552 & 16·26·32 & 17·27·29 & 36·52·59 & 44·60·68 & \textbf{81·92·94} \\
& T1600 & T1600 & 12·21·25 & 15·21·22 & 26·43·47 & 37·55·63 & \textbf{85·90·92} \\
& T1604 & T1604 & 14·24·29 & 29·43·50 & 38·56·64 & 44·69·74 & \textbf{88·96·97} \\
\midrule
\multirow{3}{*}{\shortstack[l]{Tibetan Verse->\\Tibetan Commentary\\ Retrieval}}
& Text 1 & Text 1 & 11·23·24 & 19·36·37 & 13·26·3 & 14·24·3 & \textbf{31·90·93} \\
& Text 2 & Text 2 & 10·21·25 & 14·21·25 & 19·41·48 & 24·49·57 & \textbf{32·76·83} \\
& Text 3 & Text 3 & 04·06·08 & 06·09·11 & 05·09·12 & 05·09·12 & \textbf{15·29·36} \\
\midrule
\multirow{3}{*}{\shortstack[l]{Sanskrit Verse->\\Tibetan Commentary\\ Retrieval}}
& Text 1  & Text 1 & 11·23·24 & 02·05·07 & \textbf{15}·33·37 & 03·07·14 & 14·\textbf{43·46} \\
& Text 2 & Text 2 & 10·21·25 & 02·03·03 & 03·06·08 & 20·41·48 & \textbf{30·68·77} \\
& Text 3 & Text 3 & 04·06·08 & 00·00·03 & 03·12·16 & 03·06·09 & \textbf{10·20·26} \\
\midrule
\multirow{4}{*}{\shortstack[l]{Cross-lingual\\QA\\Retrieval}}
& English Q & Pāli A & 02·04·05 & 01·03·05 & 13·23·27 & 21·34·41 & \textbf{36·55·62} \\
& English Q & Sanskrit A & 03·06·08 & 03·07·09 & 28·39·45 & 46·60·64 & \textbf{56·72·76} \\
& English Q & Tibetan A & 02·04·06 & 05·11·14 & 15·26·31 & 28·43·49 & \textbf{49·64·69} \\
& English Q & Chinese A & 02·03·05 & 06·10·13 & 12·21·26 & 25·39·44 & \textbf{57·72·78} \\
\bottomrule
\end{tabular}
\caption{Evaluation of different retrieval strategies for various retrieval tasks. All results are reported in P@1·P@5·P@10 accuracy. BM25, FastText, and BGE use machine translation into English as pivotal to enable the crosslingual mapping, while LaBSE and Gemma 2 Mitra Embed use the native language data directly.}
\label{tab:retrieval-results}
\end{table*}

\section{Ablation Study} \label{section:ablation-study}
In order to understand how \texttt{Gemma 2 MITRA-E} performs in comparison to other models in strictly monolingual retrieval settings, i.e., locating a Sanskrit commentary based on a Sanskrit verse without pivoting through English, we conduct an ablation study where we use only monolingual data of about 100k sentences per language as search space. We compare \texttt{Gemma 2 MITRA-E} against three different approaches: \texttt{BM25}, \texttt{FastText}, and \texttt{BGE-M3}. For \texttt{BM25} and \texttt{FastText}, we apply word segmentation and lemmatization in the case of Sanskrit with the model presented in~\citet{nehrdich2024}. For Chinese, we split after each character, effectively treating individual characters as words. For \texttt{BGE-M3} and \texttt{Gemma 2 MITRA-E}, we use the raw, unprocessed original sentences as input.\\
The results show that even in this setting, \texttt{Gemma 2 MITRA-E} significantly outperforms all other approaches, the only exception being the Sanskrit text MSABh, where the P@1 retrieval precision of \texttt{BGE-M3} is higher. In this monolingual setting, \texttt{BM25} outperforms \texttt{FastText} for Buddhist Chinese, and it matches the performance of \texttt{BGE-M3} closely. This shows that individual Chinese characters are very efficient signal for sparse retrieval methods. For Sanskrit, on the other hand, \texttt{FastText} shows significant performance advantages over \texttt{BM25}. We assume that even after lemmatization, remaining word segmentation ambiguities and the rich derivational morphology of Sanskrit, where new words can be derived via suffixes or prefixes from existing ones, and nouns can be derived from verbs and vice versa with similar but not completely identical lemmas, yield better performance for the subword-aware FastText algorithm compared to solely word-based BM25 retrieval. In the case of Sanskrit, \texttt{FastText} also outperforms \texttt{BGE-M3} in 4 out of 6 texts. All in all, the results of our ablation study show that \texttt{Gemma 2 MITRA-E} also performs excellently in a monolingual setting in comparison with other monolingual retrieval techniques. Since it can operate directly on the input string without additional preparation steps such as word segmentation and lemmatization, which are not trivial in the case of Sanskrit, it also allows for a less complex and more unified retrieval pipeline in the monolingual setting as well. 
\begin{table}[t]
\centering
\small
\setlength{\tabcolsep}{2pt}
\begin{tabular}{l|c|c|c|c}
\toprule
\textbf{Text} & \textbf{BM25} & \textbf{FastText} & \textbf{BGE M3} & \textbf{\shortstack{Gemma 2\\MITRA-E}} \\
\midrule
BGh & 28·46·54 & 63·70·75 & 69·78·80 & \textbf{90·95·96} \\
MSABh & 16·29·36 & 21·51·57 & \textbf{29}·53·58 & 14·\textbf{64·70}\\
DrāhŚS & 05·12·16 & 26·34·38 & 10·16·18 & \textbf{37·50·54}\\
LāṭŚS & 25·41·46 & 65·76·78 & 47·60·65 & \textbf{77·89·92} \\
ŚāṅkhŚS & 11·21·25 & 39·51·54 & 24·36·39 & \textbf{50·65·69}\\
JaimŚS & 10·20·54 & 28·38·43 & 20·25·30 & \textbf{35·51·57}\\
\midrule
T1552 & 61·79·84 & 11·29·35 & 68·83·86 & \textbf{81·92·94} \\
T1600 & 59·73·75 & 25·43·52 & 59·73·75 & \textbf{85·90·92} \\
T1604 & 71·88·91 & 20·41·49 & 69·84·88 & \textbf{89·96·97}\\
\bottomrule
\end{tabular}
\caption{Evaluation of retrieval strategies for monolingual specialized retrieval tasks without pivoting through English machine translation. Results are reported in P@1·P@5·P@10 accuracy scores.}
\label{tab:monolingual-results}
\end{table}

\section{Conclusion} \label{section:conclusion}
We presented MITRA, a complete pipeline that (1) mines noisy multilingual corpora for sentence-level parallels via an MT-pivot + filter/sentence align scheme, (2) compiles the 1.74M-pair \texttt{MITRA-parallel} corpus (89\% perfect/mostly-correct alignments in a manual audit), and (3) continuously pretrains and task-fine-tunes a 9B-parameter base LLM, \texttt{Gemma-2-MITRA}.\\
The machine translation fine-tuned model \texttt{Gemma-2-MITRA-MT} sets a new open-model state-of-the-art on Sanskrit, Pāli, Tibetan and Buddhist Chinese-to-English translation, outperforming the much larger Gemma-3-27B by at least +15 GEMBA on average and the best previous domain-specific model for Buddhist Chinese by +15 GEMBA. Its retrieval sibling, \texttt{Gemma-2-MITRA-E}, beats \texttt{LaBSE} and \texttt{BGE-M3} on our seven-task evaluation.\\ 
Beyond NLP tasks, these resources can be of substantial help for philologists: locating a Sanskrit--Chinese parallel or a relevant commentary passage now takes seconds via a search system powered by \texttt{Gemma-2-MITRA-E}. Because the mining recipe is language-agnostic, it can be ported to other historical traditions given a sufficiently performing MT and sentence alignment model.\\
All code, models, evaluation scripts, and the semantic retrieval evaluation dataset are released under open licenses, providing a reproducible testbed for future research. In future work, we hope to expand the \texttt{Gemma-2-MITRA} model to include relevant resources in other languages of the Buddhist tradition such as Tocharian or classical Japanese and modern research languages such as Japanese, French, or modern Chinese as well. Another vector of future work is the distillation of the semantic embedding model \texttt{Gemma-2-MITRA-E}, since its high parameter count and high dimensionality of the vectors currently make it challenging to apply on large corpora. 
\section{Limitations} \label{section:limitations}
The semantic similarity benchmark currently does not involve any cross-lingual retrieval tasks between Pāli and other ancient Buddhist languages, which is a noteworthy limitation. Adding these data points requires significant manual data annotation. While the current evaluation benchmark is distributed openly, the data used for the machine translation benchmark cannot be made accessible since we do not hold the rights to these works.\\
While our paper focuses on Sanskrit, Pāli, Tibetan, and Chinese, other languages of the Buddhist traditions such as classical Japanese, Tocharian, Mongolian, as well as modern material in Japanese, Korean, modern Chinese, French, and more are not yet taken into account.
\bibliography{anthology}

\appendix

\end{document}